\documentclass{article}
\usepackage{spconf,amsmath,graphicx,hyperref,tabularx}
\usepackage{array}
\usepackage{booktabs}
\usepackage{cite}
\usepackage{hyperref}
\usepackage{colortbl}  
\usepackage{xcolor} 
\usepackage{amssymb}
\usepackage{nicematrix}
\usepackage{booktabs}
\usepackage{multirow}
\usepackage{siunitx} 
\definecolor{mygray}{gray}{.9}

\title{QUSR: Quality-Aware and Uncertainty-Guided Image Super-Resolution Diffusion Model}
\name{Junjie Yin, Jiaju Li, Hanfa Xing$^{\dagger}$\thanks{$^{\dagger}$Corresponding author.}
\thanks{This work was supported by the National Natural Science Foundation of China (General Program) under Grant 42271470.}}
\address{School of BeiDou Research Institute, South China Normal University, China}
\begin{document}
\ninept
\maketitle
\begin{abstract}
Diffusion-based image super-resolution (ISR) has shown strong potential, but it still struggles in real-world scenarios where degradations are unknown and spatially non-uniform, often resulting in lost details or visual artifacts. To address this challenge, we propose a novel super-resolution diffusion model, QUSR, which integrates a Quality-Aware Prior (QAP) with an Uncertainty-Guided Noise Generation (UNG) module. The UNG module adaptively adjusts the noise injection intensity, applying stronger perturbations to high-uncertainty regions (e.g., edges and textures) to reconstruct complex details, while minimizing noise in low-uncertainty regions (e.g., flat areas) to preserve original information. Concurrently, the QAP leverages an advanced Multimodal Large Language Model (MLLM) to generate reliable quality descriptions, providing an effective and interpretable quality prior for the restoration process. Experimental results confirm that QUSR can produce high-fidelity and high-realism images in real-world scenarios. The source code is available at~\url{https://github.com/oTvTog/QUSR}.
\end{abstract}
\section{Introduction}
\label{sec:intro}

Image restoration is a critical task that has seen significant development due to its importance to downstream tasks \cite{chen2022snowformer,chen2023msp,lin2024diffbir}. Among these, image super-resolution (ISR) aims to reconstruct a clear, high-quality (HQ) image from its degraded, low-quality (LQ) counterpart. The core challenge of this problem lies in its ill-posed nature, particularly in real-world scenarios where the degradation process is both unknown and complex. To address this issue, existing super-resolution methods have explored a variety of advanced network architectures and sophisticated degradation models. Although methods based on Generative Adversarial Networks (GANs) \cite{wang2018esrgan,wang2021real} have significantly advanced the field of SR, they still face a bottleneck in enhancing perceptual quality, especially in generating fine-grained textures, often introducing numerous visual artifacts. This is primarily because their optimization objectives tend to prioritize pixel-wise fidelity over visual realism, and models trained on synthetic data struggle to generalize to real-world images due to the inherent domain gap.

With the recent rise of diffusion models in the image generation domain, their progressive denoising mechanism has offered a new paradigm for real-world tasks such as image super-resolution and restoration \cite{chen2025teaching, wang2024exploiting, wu2024one,lin2024diffbir}. In particular, large-scale pre-trained Text-to-Image (T2I) models, exemplified by Stable Diffusion (SD), have been successfully applied to a variety of downstream tasks owing to their remarkable semantic understanding and powerful generative priors \cite{lin2025jarvisir,chen2025postercraft}. To effectively leverage the potential of these pre-trained models, many studies have begun to employ techniques like ControlNet \cite{zhang2023adding} to guide and control the generation process. This has demonstrated an exceptional capability for producing photorealistic and detail-rich images in applications like super-resolution and image restoration\cite{wang2024exploiting,lin2024diffbir,yang2024pixel,wu2024seesr,sun2025pixel,qu2024xpsr}.

Current diffusion-based SR methods face key limitations. Models like StableSR\cite{wang2024exploiting} and DiffBIR\cite{lin2024diffbir}, which condition on the low-resolution image, struggle to extract effective semantic information when the input is highly degraded. In contrast, methods such as SeeSR\cite{wu2024seesr} and PiSA-SR\cite{sun2025pixel} use external models to generate text prompts, but these semantic descriptions overlook crucial degradation information (e.g., blur, noise) necessary for accurate restoration. Although XPSR\cite{qu2024xpsr} employs a dual-prompting mechanism with MLLM to describe both content and degradation, its performance heavily depends on the accuracy of the MLLM's judgments.

To enhance super-resolution performance, and inspired by XPSR\cite{qu2024xpsr} and UPSR\cite{zhang2025uncertainty}, we identify a central challenge in existing methods: the difficulty of reconciling high-level semantic guidance with low-level spatial fidelity when utilizing priors from the LQ image. A sole reliance on high-level text prompts tends to overlook inherent image degradation, while direct feature extraction from the LQ image is often corrupted by noise and artifacts. To address the limitations of current approaches, specifically their insufficient understanding of global semantic content and quality attributes coupled with their inability to spatially adapt to local reconstruction difficulty, we propose an innovative dual-guidance framework named QUSR. The main contributions of this paper can be summarized as follows:

\begin{itemize}
  \item [$\bullet$] 
We propose a Quality-Aware Prior generated by a powerful MLLM. This prior provides a comprehensive textual description that encapsulates both the semantic content and the specific degradation attributes (e.g., clarity, noise) of the input image, thereby furnishing the model with holistic semantic guidance.
\item [$\bullet$] 
We design an uncertainty-guided noise generation mechanism within a single-step residual diffusion framework. This mechanism estimates the restoration difficulty of each image region, applying minimal noise to flat areas to preserve fidelity while injecting stronger noise into complex textures to stimulate detail synthesis. To complement this, we introduce an uncertainty loss function that relaxes the reconstruction constraint on complex regions during training, enabling the model to focus on generating plausible details.
\item [$\bullet$]
On two real-world datasets, our method demonstrates superior performance by generating images with both high fidelity and photorealism.
\end{itemize}

\section{Method}
\label{sec:format}
\begin{figure*}[!t]
\centering
\setlength{\abovecaptionskip}{-0.5cm} 
\setlength{\belowcaptionskip}{-0cm}
\includegraphics[width=0.8\textwidth]{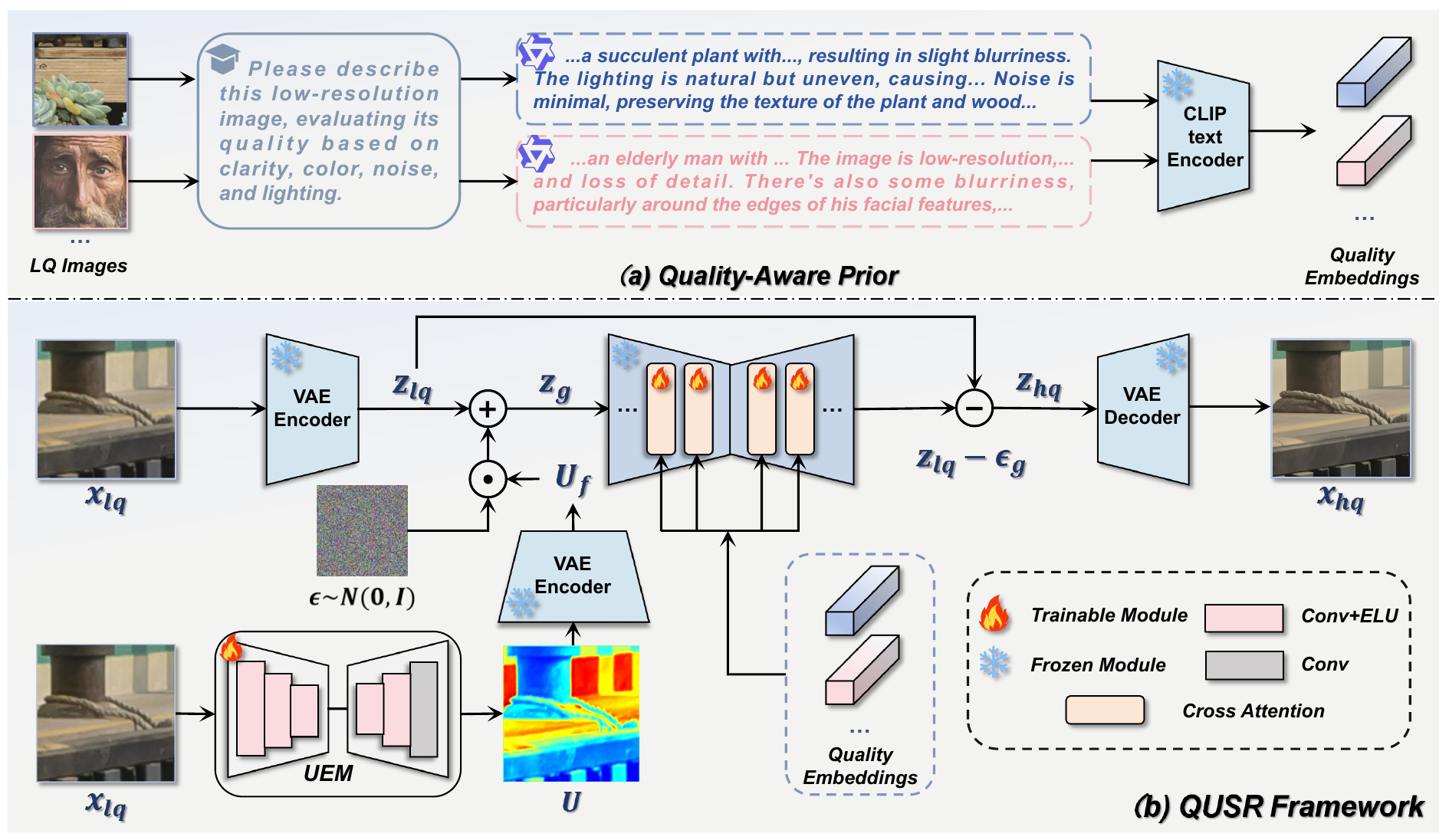}
\caption{Framework of QUSR. (a) Initially, a MLLM is employed to generate quality-aware priors, each describing the content and degradation of the corresponding LQ image. These priors are then processed by a CLIP text encoder to derive quality embeddings. (b) Subsequently,, the LQ image is processed by the QUSR framework. Its core is a single-step denoising UNet that, conditioned on the quality embeddings, predicts the noise residual ($\boldsymbol{\epsilon_g}$) from the guided latent representation ($\boldsymbol{z_g}$). Furthermore, the UEM generates an uncertainty map to construct adaptive noise perturbations, guiding the model to focus more on complex regions of the image during restoration.}
\label{f1}
\end{figure*}

We propose QUSR, a novel image super-resolution framework based on a residual diffusion model, integrating adaptive uncertainty guidance with quality-aware prompting. The overall architecture is depicted in Fig.\ref{f1}.


\subsection{Framework of QUSR}
The core backbone network of QUSR is based on the UNet denoising model from Stable Diffusion and is optimized through parameter-efficient Low-Rank Adaptation (LoRA) fine-tuning. The overall process begins with the VAE encoder \(\mathcal{E}\), which maps the input LQ image \( x_{lq} \) to a latent representation \(\boldsymbol{z_{lq}} = \mathcal{E}(x_{lq})\). Subsequently, the initial latent representation \(\boldsymbol{z_{lq}}\) is perturbed by adaptive noise generated by the uncertainty estimation module, producing a guided latent representation \(\boldsymbol{z_g}\). Then, the UNet network takes this perturbed, guided latent representation \(\boldsymbol{z_g}\) as its primary input and, guided by the quality-aware prior \(\boldsymbol{C}_q\), predicts the added noise residual \(\boldsymbol{\epsilon_g}\):
\begin{equation}
\boldsymbol{\epsilon_g} = f(\boldsymbol{z_g}, t, \boldsymbol{C}_q),
\end{equation}
where \( t \) represents the diffusion timestep (\( t =1\)), and the conditional embeddings \(\boldsymbol{C}_q\) are integrated into each layer of the UNet via a cross-attention mechanism.

The high-resolution latent representation \(\boldsymbol{z_{hq}}\) is then obtained by subtracting the predicted residual from the original latent representation:
\begin{equation}
\boldsymbol{z_{hq}} = \boldsymbol{z_{lq}} - \boldsymbol{\epsilon_g}.
\end{equation}

Finally, the final HQ image \( x_{hq} = \mathcal{D}(\boldsymbol{z_{hq}}) \) is generated through the VAE decoder \(\mathcal{D}\).

\begin{table*}[ht]
\centering
\fontsize{8pt}{10pt}\selectfont
\caption{Quantitative comparison among the state-of-the-art DM-based SR methods on real-world test datasets. The best and second-best results are highlighted in \textbf{\textcolor{red}{red}} and \textbf{\textcolor{blue}{blue}}, respectively. The symbols $\uparrow$ and $\downarrow$ denote that higher and lower values are preferable, respectively.}\label{1}
\begin{tabular*}{\textwidth}{l |l| @{\extracolsep{\fill}} S[table-format=2.2] S[table-format=1.4] S[table-format=1.4]  S[table-format=3.2] S[table-format=1.4] S[table-format=2.2] S[table-format=1.4]} 
\hline
\textbf{Datasets} & \textbf{Method} & {\textbf{PSNR}$\uparrow$} & {\textbf{SSIM}$\uparrow$} & {\textbf{LPIPS}$\downarrow$} & {\textbf{FID}$\downarrow$} & {\textbf{CLIPIQA}$\uparrow$} & {\textbf{MUSIQ}$\uparrow$} & {\textbf{MANIQA}$\uparrow$} \\
\hline 
\multirow{6}{*}{\textbf{RealSR}} 
    & StableSR\cite{wang2024exploiting} & 24.69 & 0.7052 & 0.3091 &  127.20 & 0.6195 & 65.42 & 0.6211 \\
    & SeeSR\cite{wu2024seesr} & 25.33 & 0.7273 & 0.2985 &  125.66 & 0.6594 & \textcolor{blue}{\bfseries69.37} & 0.6439 \\ 
    & SinSR\cite{wang2024sinsr} & \textcolor{red}{\bfseries26.30} & \textcolor{blue}{\bfseries0.7354}  & 0.3212 & 137.05 & 0.6204 & 60.41 & 0.5389 \\ 
    & OSEDiff\cite{wu2024one} & 25.15 & 0.7341  & \textcolor{blue}{\bfseries0.2921} & \textcolor{red}{\bfseries123.50} & 0.6693 & 69.09 & 0.6339 \\ 
    & PiSA-SR\cite{sun2025pixel} & 25.50 & \textcolor{red}{\bfseries0.7414} &  \textcolor{red}{\bfseries0.2672} & \textcolor{blue}{\bfseries124.09} & \textcolor{blue}{\bfseries0.6702} & \textcolor{red}{\bfseries70.15} & \textcolor{blue}{\bfseries0.6560} \\ 
    & Ours & \textcolor{blue}{\bfseries25.54} & 0.7289 & 0.2974 & 125.27& \textcolor{red}{\bfseries0.6824} & {69.17} & \textcolor{red}{\bfseries0.6564} \\ 
\hline 
\multirow{6}{*}{\textbf{DrealSR}} 
    & StableSR\cite{wang2024exploiting} & 28.04 & 0.7460 & 0.3354  & 147.03 & 0.6171 & 58.50 & 0.5602 \\
    & SeeSR\cite{wu2024seesr} & 28.26 & 0.7698 & 0.3197  & 149.86 & 0.6672 & 64.84 & 0.6026 \\ 
    & SinSR\cite{wang2024sinsr} & \textcolor{blue}{\bfseries28.41} & 0.7495 & 0.3741  & 177.05 & 0.6367 & 55.34 & 0.4898 \\ 
    & OSEDiff\cite{wu2024one} & 27.92 & \textcolor{blue}{\bfseries0.7835} & 0.2968 &  135.29 & 0.6963 & 64.65 & 0.5899 \\ 
    & PiSA-SR\cite{sun2025pixel} & 28.31 & 0.7804 & \textcolor{blue}{\bfseries0.2960}  & \textcolor{blue}{\bfseries130.61} & \textcolor{blue}{\bfseries0.6970} & \textcolor{blue}{\bfseries66.11} & \textcolor{blue}{\bfseries0.6156} \\ 
    & Ours & \textcolor{red}{\bfseries29.81} & \textcolor{red}{\bfseries0.8200} & \textcolor{red}{\bfseries0.2708} & \textcolor{red}{\bfseries113.87} & \textcolor{red}{\bfseries0.7082} & \textcolor{red}{\bfseries67.00} & \textcolor{red}{\bfseries0.6415} \\ 
\hline
\end{tabular*}
\end{table*}

\subsection{Quality-Aware Prior}
Leveraging multimodal pre-training and joint optimization over large-scale datasets, MLLMs exhibit exceptional capabilities in semantic understanding\cite{wu2023q,you2024depicting}. Building on this foundation, we employ Qwen2.5-VL-7B-Instruct\cite{bai2025qwen2} from Qwen to extract quality priors from LQ images, thereby enhancing the model's ability to perceive image quality and address quality issues in LQ images more effectively. We utilize the prompt instruction: \textbf{\textit{"Please describe this low-resolution image, evaluating its quality based on clarity, color, noise, and lighting."}} This approach generates descriptions that encompass key information, including the general content, overall quality, sharpness, noise level, and lighting, as demonstrated in Fig.\ref{f1}(a), where Qwen2.5-VL-7B-Instruct delivers quality descriptions aligned with human perception.

These quality descriptions are processed by the CLIP text encoder to generate quality embeddings, defined as \( \boldsymbol{C}_q = \mathcal{E}_{clip}(\boldsymbol{P}_q) \), where \( \boldsymbol{P}_q \) represents the quality prompt. These embeddings serve as conditional inputs to the cross-attention layers of the UNet network. The specific formula is as follows:
\begin{equation}\label{eq1}
\boldsymbol{F}_m^\prime = \text{softmax} \left( \frac{Q(\boldsymbol{F}_m) \cdot K(\boldsymbol{C}_q)^\mathrm{T}}{\sqrt{L}} \right) \cdot V(\boldsymbol{C}_q).
\end{equation}
where $Q(\cdot)$, $K(\cdot)$, and $V(\cdot)$ denote the query, key, and value projections, respectively. $\boldsymbol{F}_m$ denotes the feature map at the $m$-th layer of the UNet network, and $\boldsymbol{F}_m^\prime$ represents the embedded output feature.

\subsection{Uncertainty-Guided Noise Generation}
Inspired by \cite{zhang2025uncertainty,chen2023sparse}, this module adaptively generates and injects noise based on the estimated uncertainty of the input image. The process consists of two main stages: Uncertainty Map Generation and Adaptive Noise Formulation.
\subsubsection{Uncertainty Map Generation}
The Uncertainty Estimation Module (UEM) utilizes a lightweight encoder-decoder architecture. The encoder comprises three $3\times3$ convolutional layers, each followed by an Exponential Linear Unit (ELU), while the decoder mirrors this structure but omits the final activation function. The LQ image $x_{lq}$ is processed by the UEM to produce an initial, raw uncertainty map $\boldsymbol{\mathcal U}$.
\begin{equation}\label{eq2}
\boldsymbol{\mathcal U} = \mathcal{D}_{uem}(\mathcal{E}_{uem}(x_{lq})),
\end{equation}
The uncertainty map $\boldsymbol{\mathcal U}$ represents the pixel-wise aleatoric error-scale, underpinning both the spatially variant noise schedule and the heteroscedastic reconstruction objective.

\subsubsection{Adaptive Noise Formulation}
To refine the uncertainty representation, the initial map $\boldsymbol{\mathcal U}$ is first projected into the latent space by the main VAE encoder $\mathcal{E}$ and scaled by a factor $k$, yielding the latent uncertainty $\boldsymbol{\mathcal U_l} = k \cdot \mathcal{E}(\boldsymbol{\mathcal U})$. A minimum noise constraint $m \in [0, 1]$ is then applied to produce the final uncertainty $\boldsymbol{\mathcal U_f}$:
\begin{equation}\label{eq3}
\boldsymbol{\mathcal U_f} = m + (1 - m) \cdot \boldsymbol{\mathcal U_l},
\end{equation}
This constraint prevents unstable perturbations that could arise from excessively low noise. Based on this final uncertainty, the noise standard deviation $\sigma_\epsilon$ is computed. To ensure numerical stability, a small constant $\delta > 0$ is added:
\begin{equation}\label{eq4}
\sigma_\epsilon = \sqrt{|\boldsymbol{\mathcal U_f}| + \delta},
\end{equation}
This mechanism ensures that high-uncertainty regions (e.g., edges, textures) receive stronger noise perturbations to facilitate the reconstruction of complex details, while low-uncertainty regions (e.g., flat areas) receive minimal noise, thus preserving information from the original input.

Finally, the guided latent representation $\boldsymbol{z_g}$ is formed by adding scaled Gaussian noise to the initial latent representation $\boldsymbol{z_{lq}}$:
\begin{equation}\label{eq5}
\boldsymbol{z_g} = \boldsymbol{z_{lq}} + \boldsymbol{\epsilon} \cdot \sigma_\epsilon \cdot p.
\end{equation}
where $\boldsymbol{\epsilon} \sim \mathcal{N}(0, 1)$ is a standard Gaussian noise tensor and $p$ is a scalar representing the perturbation strength. This entire process optimizes the trade-off between information preservation and detail reconstruction during super-resolution.

\subsection{Loss Function}
To guide the effective training of our model, we design a composite loss function $\mathcal{L}$. This function is composed of four weighted loss terms, designed to synergistically optimize the quality of the generated images across multiple dimensions. The total objective function is defined as:
\begin{equation}
\mathcal{L} = \lambda_1 \mathcal{L}_{2} + \lambda_2 \mathcal{L}_{lpips} + \lambda_3 \mathcal{L}_{csd} + \lambda_4 \mathcal{L}_{un},
\end{equation}
where $\lambda_1, \lambda_2, \lambda_3,$ and $\lambda_4$ are the hyperparameters that weight the contribution of each loss term. The L$2$ loss (\(\mathcal{L}_{2}\)) penalizes pixel-level differences to ensure content fidelity between generated and ground truth images. The LPIPS\cite{zhang2018unreasonable} loss (\(\mathcal{L}_{lpips}\)) uses the LPIPS metric to enhance visual realism by assessing perceptual similarity in the deep feature space. The Classifier Score Distillation (CSD)\cite{ho2022classifier} loss (\(\mathcal{L}_{csd}\)) employs a pre-trained Stable Diffusion model as an implicit classifier, leveraging Classifier-Free Guidance (CFG) to extract semantic gradients, ensuring super-resolved results align visually and semantically with quality-aware prompts.

Furthermore, we introduce an uncertainty loss ($\mathcal{L}_{un}$) that leverages the estimated uncertainty information to guide the optimization process. This loss guides the model to maintain high reconstruction fidelity in regions of low uncertainty (e.g., smooth backgrounds) while permitting some reconstruction error in regions of high uncertainty (e.g., complex textures). Its formula is defined as:
\begin{equation}
\mathcal{L}_{un} = \mathcal{L}_{1}(x_{hq} \cdot \exp(-\boldsymbol{\mathcal U}_n), x_{gt} \cdot \exp(-\boldsymbol{\mathcal U}_n)) + \alpha \cdot \text{mean}(\boldsymbol{\mathcal U}_n).
\end{equation}
where $\mathcal{L}_{1}$ denotes the L1 loss function. The weight term $\exp(-\boldsymbol{\mathcal U}_n)$ is derived directly from the normalized uncertainty map $\boldsymbol{\mathcal U}_n$. This strategy prioritizes the reconstruction fidelity of low-uncertainty regions. The second term is a regularization term controlled by the hyperparameter $\alpha$, which constrains the overall distribution of uncertainty to prevent the model from producing trivially high uncertainty across the entire image.

\section{Experiments}
\label{sec:pagestyle}
\begin{figure*}[!t]
\centering
\setlength{\abovecaptionskip}{0.1cm} 
\setlength{\belowcaptionskip}{-0cm}
\includegraphics[width=0.7\textwidth]{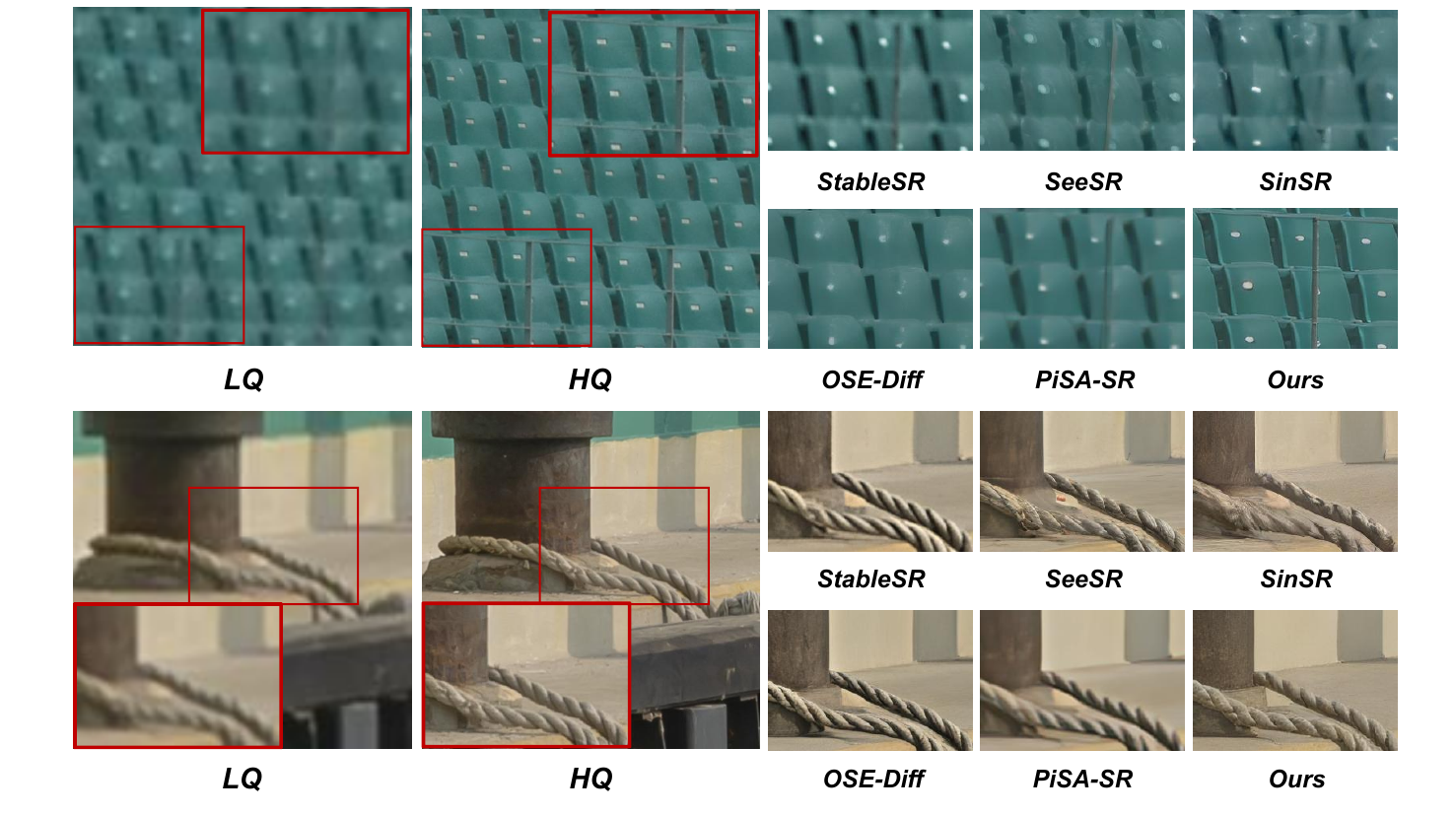}
\caption{Visual comparison of our method (QUSR) with SOTA methods on the RealSR and DRealSR datasets.}
\label{f2}
\end{figure*}
\textbf{Datasets and Metrics.} Following SeeSR \cite{wu2024seesr}, we employ the LSDIR \cite{li2023lsdir} dataset along with the initial $10k$ images from the FFHQ \cite{karras2019style} dataset for training, with LQ-HQ training pairs of $512\times512$ resolution pre-generated through the RealESRGAN\cite{wang2021real} degradation pipeline. The test dataset comprises center-cropped LQ-HQ image pairs sourced from the RealSR\cite{cai2019toward} and DRealSR\cite{wei2020component} datasets. All LQ images in the test sets are resized to $128\times128$, while the corresponding HQ images are $512\times512$. Evaluation metrics include reference-based measures: PSNR, SSIM, LPIPS, DISTS, and FID, as well as no-reference measures: CLIPIQA, MUSIQ, and MANIQA.

\noindent\textbf{Implementation Details.} 
The QUSR model is built upon the Stable Diffusion $2.1$ framework and is designed for $\times4$ image super-resolution tasks. Training is conducted on four $24$GB NVIDIA RTX $3090$ GPUs, using the Adam optimizer with an initial learning rate of $3 \times 10^{-5}$ and a batch size of $4$, totaling $15$K iterations. The LoRA rank is set to $4$. For the loss function, the weight coefficients are \(\lambda_1 = 0.5\), \(\lambda_2 = 2\), \(\lambda_3 = 2\), and \(\lambda_4 = 0.3\).

\noindent\textbf{Quantitative Analysis.} 
In this section, we quantitatively evaluate our proposed QUSR against other leading diffusion-based methods\cite{wang2024exploiting,wu2024seesr,wang2024sinsr,wu2024one,sun2025pixel} on two real-world datasets: RealSR\cite{cai2019toward} and DRealSR\cite{wei2020component}. To ensure a fair comparison, all competing methods are tested using their officially released pre-trained models, and all metric scores are sourced from PiSA-SR\cite{sun2025pixel}. As shown in Table \ref{1}, our method demonstrates superior performance on both test benchmarks. Notably, on the DRealSR dataset, QUSR achieves state-of-the-art (SOTA) results across all metrics. For instance, compared to the second-best performing method, it reduces the FID score by $16.74$ while increasing the MUSIQ score by $0.89$, highlighting its superiority in generating images with both high fidelity and perceptual quality.

\noindent\textbf{Visual Comparison.} 
Fig.\ref{f2}. illustrates the visual comparison between QUSR and other state-of-the-art super-resolution methods on real-world images. As can be seen, existing methods often struggle to generate accurate and realistic details, particularly in regions with complex edges and fine textures. In contrast, QUSR effectively mitigates this issue, producing details that are both structurally more accurate and visually more natural. Furthermore, QUSR demonstrates superior fidelity when handling dense, repetitive textures and significantly reduces visual artifacts.

\noindent\textbf{Ablation Study.} 
We validate the efficacy of two core modules: the Quality-Aware Prior (QAP) and the Uncertainty-Guided Noise generation (UNG). As presented in Table \ref{2}, we evaluate three model variants: the model without the QAP module (w/o QAP), without the UNG module (w/o UNG), and without both modules, which serves as our baseline.
\begin{table}[htbp]
\centering
\setlength{\tabcolsep}{6.5pt} 
\fontsize{7pt}{8pt}\selectfont
\caption{Ablation study of Quality-Aware Prior and Uncertainty.}
\begin{tabular*}{\columnwidth}{
    l
    S[table-format=2.2]  
    S[table-format=1.4]
    S[table-format=1.4]  
    S[table-format=2.2]  
    S[table-format=1.4] 
}
\toprule
\textbf{Method} & {\textbf{PSNR}$\uparrow$} & {\textbf{SSIM}$\uparrow$} & {\textbf{CLIPIQA}$\uparrow$} & {\textbf{MUSIQ}$\uparrow$} & {\textbf{MANIQA}$\uparrow$} \\
\midrule
w/o QAP& {\bfseries 30.19} & {\bfseries 0.8206} & 0.6853 & 66.63 & 0.6318 \\
w/o UNG & 29.74  & 0.8179& 0.6906 & 66.58 & 0.6392 \\
Baseline & 29.05& 0.8071& 0.6745 & 65.54 & 0.6268 \\
\midrule 
\rowcolor{mygray}
Ours& 29.81& 0.8200& {\bfseries 0.7082} & {\bfseries 67.00} & {\bfseries 0.6415}\\
\bottomrule
\end{tabular*}
\label{2}
\end{table}

Removing the Quality-Aware Prior (w/o QAP) leads to a slight increase in fidelity-oriented metrics such as PSNR/SSIM, but causes a significant drop in perception-oriented no-reference metrics like MUSIQ. This result validates that the global semantic and degradation priors provided by the QAP module are crucial for guiding the model to generate realistic details that align with human perception.

Removing the Uncertainty-Guided Noise generation (w/o UNG), in turn, results in a comprehensive decline across all metrics. This clearly demonstrates that the adaptive noise injection strategy of the UNG module plays a decisive role in the fine-grained reconstruction of complex textures and in preventing the image from being over-smoothed.

\section{Conclusion}
\label{sec:typestyle}
This paper introduces QUSR, a diffusion-based framework for image super-resolution. We leverage Qwen2.5-VL to generate a Quality-Aware Prior, providing comprehensive descriptions of content and degradation to offer effective global semantic guidance. Concurrently, an Uncertainty-Guided Noise module is proposed to adaptively modulate noise injection intensity, prioritizing the reconstruction of complex textures while preserving flat regions. By integrating these two mechanisms, QUSR balances high-level semantic guidance with low-level spatial fidelity, achieving superior performance and photorealism in real-world tasks.


\vfill\pagebreak

\bibliographystyle{IEEEbib}
\bibliography{refs}

\end{document}